# Adversarial Training for Adverse Conditions: Robust Metric Localisation using Appearance Transfer

Horia Porav, Will Maddern and Paul Newman

*Abstract*— We present a method of improving visual place recognition and metric localisation under very strong appearance change. We learn an invertable generator that can transform the conditions of images, e.g. from day to night, summer to winter etc. This image transforming filter is explicitly designed to aid and abet feature-matching using a new loss based on SURF detector and dense descriptor maps. A network is trained to output synthetic images optimised for feature matching given only an input RGB image, and these generated images are used to localize the robot against a previously built map using traditional sparse matching approaches. We benchmark our results using multiple traversals of the Oxford RobotCar Dataset over a year-long period, using one traversal as a map and the other to localise. We show that this method significantly improves place recognition and localisation under changing and adverse conditions, while reducing the number of mapping runs needed to successfully achieve reliable localisation.

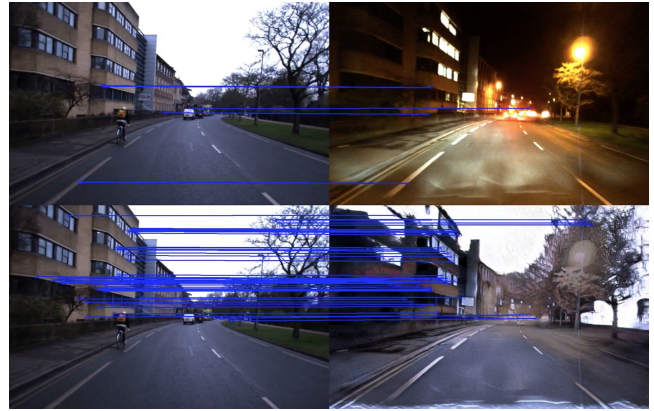

Fig. 1. Improved feature-based localisation using synthesized images. The top row shows point feature matches between a day map keyframe (left) and the live camera night image (right). The bottom row shows the improved matches between the day map keyframe (left) and the synthetic day image (right) obtained by transforming the live camera night image. The key to our approach is explicitly accounting for feature detection and descriptor extraction using differentiable components of the synthetic image pipeline. Our system drastically reduces the amount of mapping needed to achieve successful localisation under challenging adverse conditions.

## I. INTRODUCTION

In this paper we present a way of improving the performance of camera-only metric localization in environments with large changes in appearance, (e.g. day-night) using images synthesized by a CNN, and introduce a new loss based on SURF detector and descriptor maps that improves the feature matching between the synthesized and real images.

Keypoint feature matching (such as SURF, SIFT, BRIEF or ORB) represents the leading approach in multi-map visual localisation systems, as used in [1], [2], [3], [4], [5] and many other commercial systems. While this works well when matching scenes with similar lighting conditions, performance quickly breaks down when the appearance of the scenes differs due to changes in illumination levels, or seasonal differences. Attempting to use standard descriptors will likely lead to poor or impoverished localisation. In order for these point-feature based methods to work, the mapping procedure would typically need to be performed for each category of appearance (sunny, snow, dawn, dusk, rain, night etc). While most approaches so far have attempted to improve the feature detectors and descriptors, we propose to instead modify the input images. In this way, the modified images can be used with any existing system based on feature-matching, without changing it. This paper builds on previous work by [6], [3] and [7], and is motivated by two drawbacks of the current approaches :

1) Mapping for all conditions is time consuming, expensive and may well be impossible in certain cases.
2) The quality of data collected during adverse conditions can be worse, leading to lower accuracy in the generated maps.

Authors are from the Oxford Robotics Institute, University of Oxford, UK. {horia,wm,pnewman}@robots.ox.ac.uk

The key contribution of our work is to incorporate a differentiable feature detector and descriptor pipeline as part of an adversarially-trained network designed to synthesise images of the same location under different conditions. Building on the work of [7], we train a CNN to output synthetic images in two steps:

- We select a complete source-condition traversal and a small portion of a target-condition traversal and, without any alignment of the images, train a pair of image synthesizers using a cycle-consistency loss, discriminator loss and feature detector and descriptor losses.
- Afterwards, using metric 6DoF groundtruth, we select a small number of well aligned target- and source-condition image pairs, and separately fine-tune the image synthesizers using an L1 loss on the feature detector and descriptor outputs.

We expand on suggestions by [8] and [9] to speed up descriptor computations, and take inspiration from [10], who introduce differentiable HOG descriptors and use it in a pipeline to improve rendered images. We describe an algorithm for quickly and efficiently computing dense, per-pixel differentiable descriptors, and use it to compute descriptor losses alongside the reconstruction and discriminator losses. The synthetic images are then used in a stereo localisation pipeline based on [3]. This pipeline first performs place recognition, outputting candidate frames that are most likely to be from the same place as the robot's live frame, and subsequently, uses keypoint feature matching to compute a metric pose between the live frame and the frame(s) retrieved during place recognition. We benchmark our localisation results against an RTK-GPS ground truth from the Oxford Robotcar Dataset introduced by [11] and present the results in section V.

## II. RELATED WORK

**Topological Localisation:** Synthetic images have seen use in topological localisation, with [12] showing that point-of-view makes a large difference when matching across images with large changes in appearance, and subsequently using Google Streetview panoramas to synthesize images that match the point of view of a query image. Co-visibility graphs for CNN features were used by [13], boosting the invariance to viewpoint changes by adding features from neighbouring views to the image that is currently queried, essentially creating 'in-between' views. In [14], features that co-occur for each image taken at different times of the day are combined into a unique representation that contains features identifiable in any image from that point of view, irrespective of illumination conditions.

Changes in an image across seasons were predicted by [15] using vocabularies of superpixels and used for place recognition. How a winter scene would look in the summer (and vice-versa) is predicted by first partitioning the image into superpixels, associating each superpixel with an entry in the winter vocabulary and then replacing the superpixel with its counterpart from the summer vocabulary by using a dictionary that maps between vocabularies. However, these synthetic views suffer from poor alignment or degradation of point features, meaning that metric localisation cannot be achieved.

**Metric Localisation:** In the context of metric localisation, [16] refine the multi-experience approach by introducing a step that creates a 'summary map' from multi-experience maps during online localisation, using a framework of sampling policies and ranking functions. However, their approach does not mitigate the need to perform mapping under various conditions. In [17], authors generate different views of the images to be matched using affine transforms and apply SIFT to these generated views, showing an increased robustness to changes in viewpoint, but without addressing changes in appearance.

High dynamic range image matching in the context of visual odometry is improved in [18] by training an LSTM deep neural network to produce temporally consistent enhanced representations of the images. However, visual odometry works with images that are temporally close with few lighting or appearance changes.

**Appearance Transfer:** Synthetic image techniques have been used for other tasks, but show promise in localisation contexts: [19] use local affine transforms in a color transfer technique to 'hallucinate' the appearance of an image at different times of day. In [20], images are decomposed into style and content representations using a VGG-19 network, followed by synthesization, starting from a white noise image and using gradient descent, of an image that matches the content representation of the input image and the style representation of the target image.

Closer to our approach, [7] demonstrate unsupervised image-to-image translation by training a pair of CNN generators, $G$ that maps $X$ to $Y$, and $F$ that maps $Y$ to $X$, and apply a 'cycle consistency' L1 loss between $X$ and $F(G(X))$ along with discriminator losses $L_G$ on the output of $G$ and $L_F$ on the output of $F$. Similarly, [21] train a pair of VAE-GANs for image to image translation using unaligned images, but do not use a cycle-consistency loss, instead choosing to partially share high-level layer weights and to share the latent-space encodings between the VAEs. The authors train on different domain transformation pairs, including semantic-labels to natural images and day images to night images, but do not demonstrate any topological or metric localisation applications. We believe we are the first to directly address metric localisation using an appearance transfer approach.

## III. LEARNING TO GENERATE IMAGES

We base our detection and matching pipeline on the well-known SURF feature [22], and employ a 2-stage training strategy. First, we use a cycle-consistency architecture similar to [7], to train a generator to transform an input source image into a synthetic image with a target condition. This synthetic image is subsequently transformed by a second generator back into a reconstructed image that has the initial condition, with the process being repeated in the reverse direction. We then fine-tune our image generators independently using a well-aligned subset of the dataset.

In the first stage, two generators, $G_{\text{AB}}$, which transforms condition A into condition B, and $G_{\text{BA}}$, which transforms condition B into condition A, are trained using a collection of unpaired source and target images. A discriminator loss is applied on the synthetic images, and an L1 loss is applied between the reconstructed images and the input images. Additionally, we compute SURF detector response maps on the reconstructed and input images and apply an L1 loss between them, and similarly compute dense per-pixel SURF descriptor maps on the reconstructed and input images and apply an L1 loss between them; these methods are further described in III-A and III-B. The architecture of the first stage is shown in Fig. 2.

In the second stage, $G_{\text{AB}}$ and $G_{\text{BA}}$ are separately trained using a small dataset of aligned day and night images. The use of pixel-aligned images allows the generators to learn certain feature transformations that might have been uncaptured by the unsupervised method used in the first stage, which only learns to align the image distributions without any explicit pixel-wise mapping. This time, we apply the L1 loss between SURF detector response maps computed on aligned target images and synthetic images, and between dense descriptor response maps computed on aligned target and synthetic images. The architecture of the second stage is shown in Fig. 3.

The generator architecture is based on UResNet [23], which combines a UNet [24] with residual (ResNet) [25]

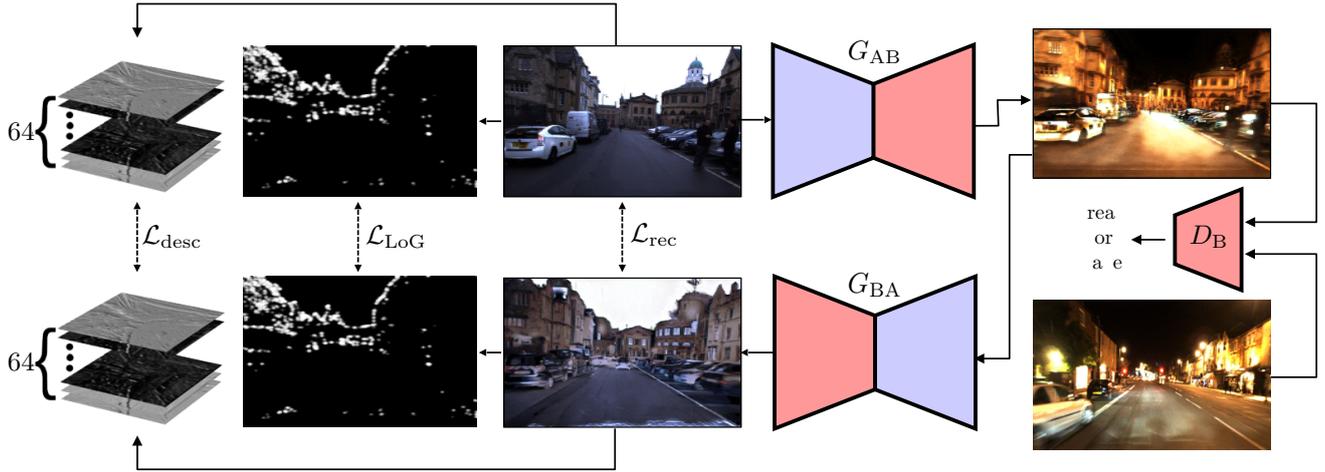

Fig. 2. Cycle-consistency architecture. This architecture is employed in stage 1 of the training process, and trains a pair of generators to transfer appearance from source to target images, and vice-versa, without requiring registration of images.

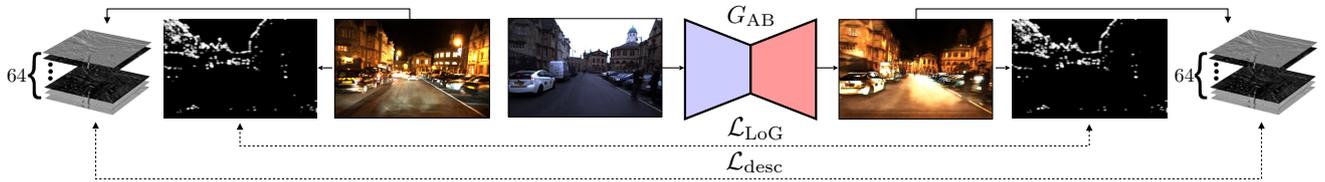

Fig. 3. Fine-tuning architecture. This architecture is employed in stage 2 of the training process on a well-aligned subset of the data in order to minimise the difference between feature detector and descriptor layers between different conditions. Note: discriminator is omitted from figure for compactness.

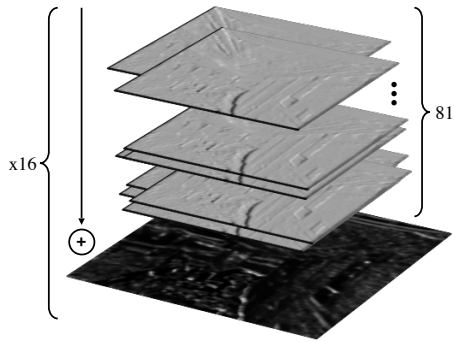

Fig. 4. HAAR response stack. For each of the 16 neighbourhood regions that compose a SURF descriptor for a specific scale, and for each of the X and Y directions, we stack 81 copies of the image convolved with the corresponding X- or Y-direction HAAR box filter, offset them according to a precomputed look-up table and multiply each response with a precomputed weight. We separately sum, along the stacking axis, the response maps and the absolute values of the response maps, for each direction, yielding 4x16 stacked matrices which represent the dense SURF descriptor.

modules. The internal architecture of the generator is shown in Fig. 5.

The discriminator architecture is a CNN with 5 layers. The first 4 layers are comprised of a convolution operation followed by instance normalisation and leaky ReLu, and the last layer is a convolution operation which outputs a $H/8 \times W/8$ map classifying the receptive field in the image space as real or fake, where $H$ and $W$ represent the height and width of the input image. Examples of generator results for both $G_{AB}$ and $G_{BA}$ are shown in Fig. 6 for a range of different condition pairs.

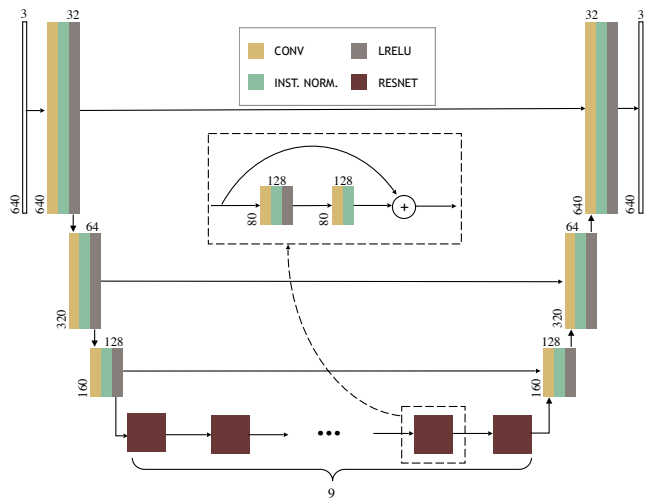

Fig. 5. Generator internal architecture. We employ 3 down-convolutional layers with stride 2, 9 ResNet blocks and 3 up-convolutional layers with fractional stride $1/2$, with skip connections between corresponding down- and up-convolutional layers. Each convolutional layer consists of a convolution operation, followed by instance normalisation and leaky ReLU. Each ResNet block consists of a convolution, followed by instance normalisation, leaky ReLU, a second convolution, instance normalisation and addition of the original block input to the resulting output.

### A. SURF Detector Response Map

The SURF detector response map is obtained using a convolutional version of the original method of approximating the determinants of Hessians described in [22]. For each scale we generate three box filters to approximate the second-order derivative of the gaussian $\frac{\partial^2}{\partial x^2}g(\sigma)$, $\frac{\partial^2}{\partial y^2}g(\sigma)$, $\frac{\partial^2}{\partial x \partial y}g(\sigma)$ on the X,Y and diagonal directions respectively. We convolve these filters with the image $I$, yielding the response maps $L_{\text{xx}}(\sigma)$, $L_{\text{yy}}(\sigma)$ and $L_{\text{xy}}(\sigma)$.

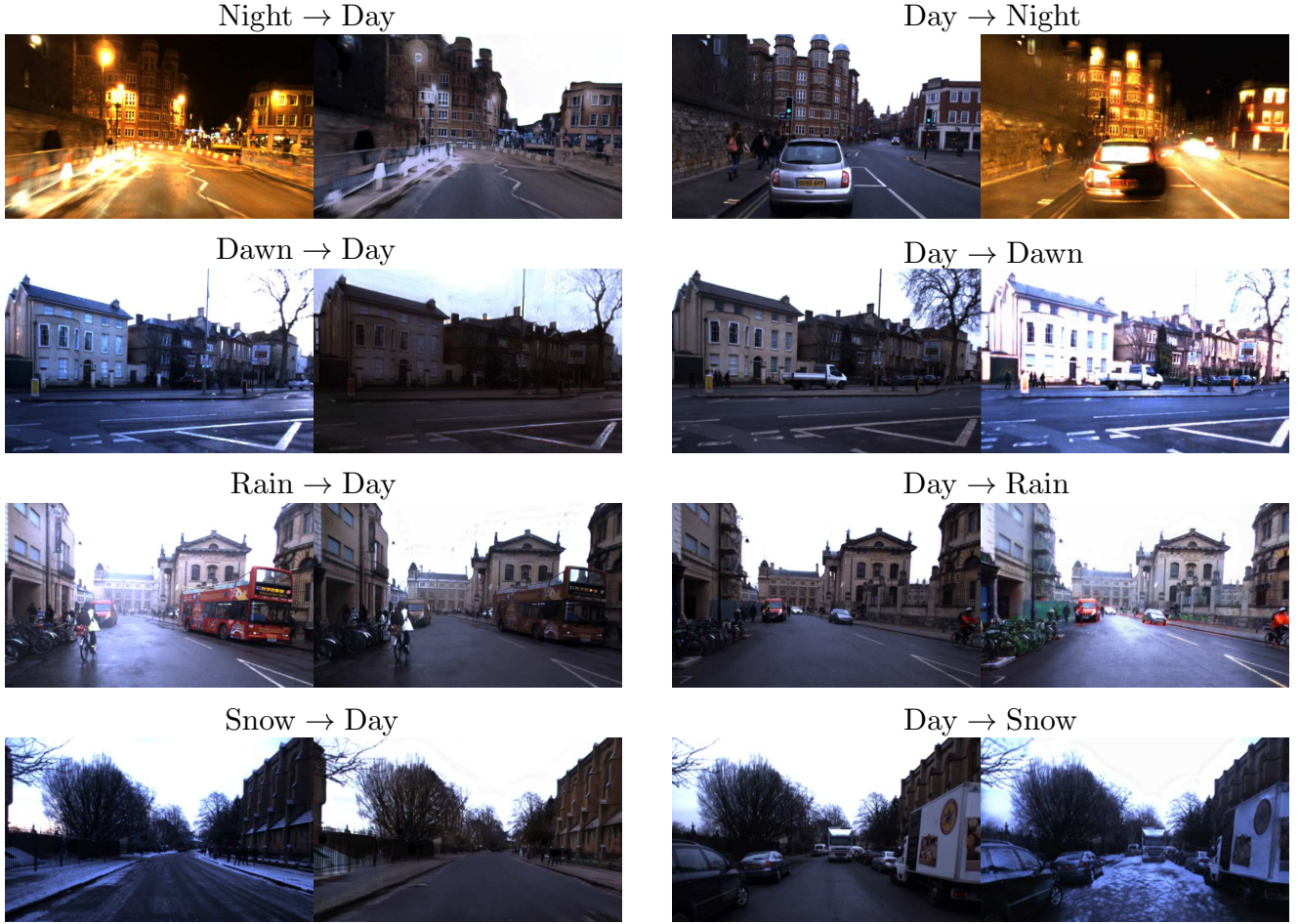

Fig. 6. Example appearance transfer between different conditions using the cycle-consistency generators $G_{\text{AB}}$ and $G_{\text{BA}}$. Each condition pair (Day/Night, Day/Dawn, Day/Rain, Day/Snow) produces a pair of generators, allowing us to perform bidirectional appearance transfer trained using only unordered collections of images from each condition.

Using the Hadamard product, the matrix of approximations of the determinant of Hessians is:

$$\det(\mathcal{H}_{\text{approx}}) = L_{\text{xx}} \circ L_{\text{yy}} - 0.81 * L_{\text{xy}} \circ L_{\text{xy}} \qquad (1)$$

*B. Dense SURF Descriptors*

We adapt the methodology used in OpenSURF [26] and employ a fast, convolutional method for building dense per-pixel SURF descriptors, through which gradients can be passed. For each scale out of the $N$ chosen scales, we pre-compute:

- a look-up table for the 81 relative offsets of pixel neighbours that are used to build SURF descriptors
- an $N \times 81$-matrix for the scale-specific Gaussian weights of the 81 offsets
- a 16-length column vector for the Gaussian weights of the 16 neighbourhoods
- HAAR-like box filters for both the X and Y directions

We then convolve the input image with the HAAR box filters and store the wavelet responses. For each chosen scale we stack 81 copies of the wavelet responses and multiply them with the scale-specific Gaussian weight.

Then, for each of the 16 pixel neighbourhoods that make up a SURF descriptor we:

- offset the stacked copies along the X and Y directions according to the offset look-up table
- multiply by the neighbourhood-specific Gaussian weight
- sum along the stack direction both the raw and absolute values for the X and Y-directions respectively, yielding 4 matrices.
- element-wise multiply each matrix with its specific Gaussian neighbourhood weight LUT
- stack the 4 resulting matrices

Finally, we normalize each column of the resulting 64 layer stack of $H \times W$ size matrices, where $H$ and $W$ are the height and width of the input images. This stack represents the dense per-pixel SURF descriptor, for each scale. The stacking and summing operation is depicted in Fig. 4.

Feature detectors and descriptors for day-night matching are evaluated in [27], showing that most features are detected at small scales (< 10). Following our own experiments, we only compute SURF loss terms for the first 5 scales in order to speed-up the training process, without noticing any significant performance loss. An explanation for this could be that inside smaller pixel neighbourhoods, appearance changes between images with different conditions can be more uniform compared to larger neighbourhoods.

*C. Losses*

Similar to [7], we apply an adversarial loss through a discriminator on the output of each generator: discriminator $D_\text{B}$ on the output of generator $G_\text{AB}$, and discriminator $D_\text{A}$ on the output of generator $G_\text{BA}$. This loss is formulated as:

$$\mathcal{L}_{\text{B}_\text{adv}} = (D_\text{B}(G_\text{AB}(I_\text{A})) - 1)^2 \quad (2)$$

$$\mathcal{L}_{\text{A}_\text{adv}} = (D_\text{A}(G_\text{BA}(I_\text{B})) - 1)^2 \quad (3)$$

The adversarial objective $\mathcal{L}_\text{adv}$ becomes:

$$\mathcal{L}_\text{adv} = \mathcal{L}_{\text{B}_\text{adv}} + \mathcal{L}_{\text{A}_\text{adv}} \quad (4)$$

The discriminators are trained to minimize the following loss:

$$\mathcal{L}_{\text{B}_\text{disc}} = (D_\text{B}(I_\text{B}) - 1)^2 + (D_\text{B}(G_\text{AB}(I_\text{A})))^2 \quad (5)$$

$$\mathcal{L}_{\text{A}_\text{disc}} = (D_\text{A}(I_\text{A}) - 1)^2 + (D_\text{A}(G_\text{BA}(I_\text{B})))^2 \quad (6)$$

The discriminator objective $\mathcal{L}_\text{disc}$ becomes:

$$\mathcal{L}_\text{disc} = \mathcal{L}_{\text{B}_\text{disc}} + \mathcal{L}_{\text{A}_\text{disc}} \quad (7)$$

The cycle consistency loss [7] is applied between the input and reconstructed image, and between the SURF detector $Det(\cdot)$ and dense descriptor $Desc(\cdot)$ maps computed from these two images:

$$\mathcal{L}_\text{rec} = \|I_\text{input} - I_\text{reconstructed}\|_1 \quad (8)$$

$$\mathcal{L}_\text{LoG} = \|Det(I_\text{input}) - Det(I_\text{reconstructed})\|_1 \quad (9)$$

$$\mathcal{L}_\text{desc} = \|Desc(I_\text{input}) - Desc(I_\text{reconstructed})\|_1 \quad (10)$$

The complete generator objective $\mathcal{L}_\text{gen}$ becomes:

$$\mathcal{L}_\text{gen} = \lambda_\text{rec} * \mathcal{L}_\text{rec} + \lambda_\text{LoG} * \mathcal{L}_\text{LoG} + \lambda_\text{desc} * \mathcal{L}_\text{desc} + \lambda_\text{adv} * \mathcal{L}_\text{adv} \quad (11)$$

Each $\lambda$ term is a hyperparameter that weights the influence of each loss component. For the fine-tuning stage, where the target image is aligned with the input and synthetic images, the losses become:

$$\mathcal{L}_{\text{F}_\text{LoG}} = \|Det(I_\text{target}) - Det(I_\text{synthetic})\|_1 \quad (12)$$

$$\mathcal{L}_{\text{F}_\text{desc}} = \|Desc(I_\text{target}) - Desc(I_\text{synthetic})\|_1 \quad (13)$$

The fine-tuning objective $\mathcal{L}_\text{finetune}$ becomes:

$$\mathcal{L}_\text{finetune} = \lambda_\text{LoG} * \mathcal{L}_{\text{F}_\text{LoG}} + \lambda_\text{desc} * \mathcal{L}_{\text{F}_\text{desc}} \quad (14)$$

We wish to find the generator functions $G_\text{AB}, G_\text{BA}$ such that:

$$G_\text{AB}, G_\text{BA} = \underset{G_\text{AB}, G_\text{BA}, D_\text{B}, D_\text{A}}{\arg\min} \mathcal{L}_\text{gen} + \mathcal{L}_\text{disc} + \mathcal{L}_\text{finetune} \quad (15)$$

In the following section we describe how the network is trained to minimise the above losses.

## IV. EXPERIMENTAL SETUP

We chose 6 traversals from the Oxford RobotCar Dataset [11] collected up to 1 year apart, yielding 5 condition pairs: day-night, day-snow, day-dawn, day-sun and day-rain. For each of the traversals, the RTK-GPS ground truth was filtered and any data points with more than 25cm of translation standard deviation were discarded.

Training datasets for each condition pair were created from the entirety of the day traversal and a portion representing approximately 20% of the paired condition, to simulate a case where reasonable amounts of mapping data cannot be acquired. The remaining 80% of the paired condition was used to benchmark the performance of the synthetic images.

The well-aligned datasets used in the second training stages were created by selecting image pairs between which none or only a small viewpoint rotation existed. Image pairs with no translation or rotation misalignment were used as-is, and for those with small rotational differences the target images were affine-warped into the frame of the source images using the known poses provided by the RTK-GPS ground truth.

*A. Training*

For the cycle-consistency stage, we used a network training regimen similar to [7]. For each iteration we first trained the discriminator on a real target domain image and a synthetic image from a previous iteration with the goal of minimizing $\mathcal{L}_\text{disc}$, and then trained the generators on input images to minimize $\mathcal{L}_\text{gen}$. We used the Adam solver [28] with an initial learning rate set at 0.0002, a batch size of 1, $\lambda_\text{rec} = 8$, $\lambda_\text{det} = 2$, $\lambda_\text{desc} = 2$ and $\lambda_\text{adv} = 1$. For the fine-tuning stage, we trained on a small well-aligned subset of the dataset, minimizing $\mathcal{L}_\text{finetune}$, with the same learning parameters.

*B. Localisation*

We used the trained generators $G_\text{AB}$ to transform all of the day map frames into target-condition frames, and $G_\text{BA}$ to transform the 5 types of target-condition frames into day-condition frames. To benchmark the synthetic images in the context of localisation, we used the experience-based navigation system of [3] which implements a feature-based topological localiser [29] followed by a geometric verification stage using RANSAC [30] and a nonlinear optimisation to minimise inlier reprojection errors. In contrast to adding the synthetic frames as a separate map, it is possible to accumulate feature correspondences from real to real image matching and synthetic to real image matching, leading to more robust and accurate solutions. The generator runs at approximately 1 Hz for images with a resolution of $1280 \times 960$, and at approximately 3 Hz for images with a resolution of $640 \times 480$ on an Nvidia Titan X GPU.

## V. RESULTS

*A. Quantitative results*

We benchmark our results taking into consideration both the frequency and the quality of localisation. Table I compares the root mean squared translation and rotation errors, with respect to the RTK-GPS ground truth, and the cumulative successfully localized portion as a percentage of the distance travelled in the case of day-night localisation. Results are shown for raw images, images obtained with an RGB-only implementation of [7], and images obtained using stages 1 and 2 of our solution. The results show a significant increase in the accuracy of localisation using synthetic images generated from the stage-1 model, and a further increase in accuracy from the stage-2 fine-tuned model.

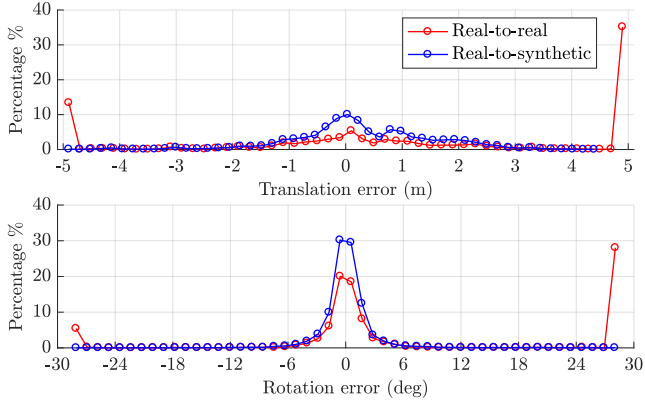
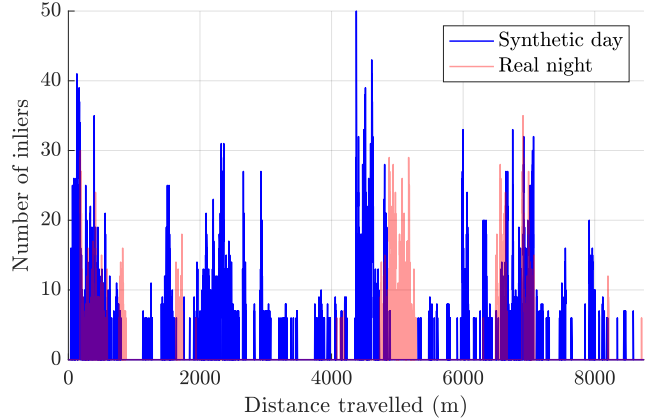

Fig. 7. Real day to real night localisation vs. Real day to synthetic day localisation : Histogram of translation and rotation error from ground truth. Translation outliers larger than 5 meters in absolute have been accumulated in the -5 and +5 meter bins. Rotation outliers larger than 30 degrees in absolute have been accumulated in the -30 and +30 degree bins.

Table II presents localisation results for a wide range of conditions transformed into day using stage-1 trained models, illustrating the performance of the method when localising relative to a single condition. In all cases the localisation rate is improved (often by a factor of 2) and the metric errors are reduced.

Fig. 7 shows the distribution of translation and rotation errors with respect to ground truth, for the raw image matching and for our best solution, in the case of day-night localisation. Overall we observe a large improvement in localisation accuracy with our solution, compared to raw images and images produced by an RGB-only implementation of [7].

Fig. 8 shows the number of match inliers as a function of the distance travelled, for the raw images and for our best solution, in the case of day-night localisation. We observe a significant increase in the number of inliers for real-to-synthetic matching, compared to real-to-real image matching.

Fig. 9 shows the probability of travelling a certain distance without localisation against the map (VO-based open-loop or dead-reckoning) when a localisation failure occurs. We observe a significant reduction in the distance to re-localisation against the map when using synthetic images generated using our solution. Surprisingly, images generated using an RGB-only implementation of [7] did not bring a large improvement in robustness.

### B. Qualitative results

We present qualitative results in a series of locations throughout Oxford where matching between raw images failed or produced a very small number of inliers. Fig. 10 presents matches between real images (top) and between real and synthetic images (bottom). Note how the learned image transformation (by construction) does a qualitatively good job reconstructing details that are described by the feature detector and descriptor, such as window frames.

Fig. 8. Inlier count for real day vs. real night localisation and real day vs. synthetic day as a function of distance travelled.

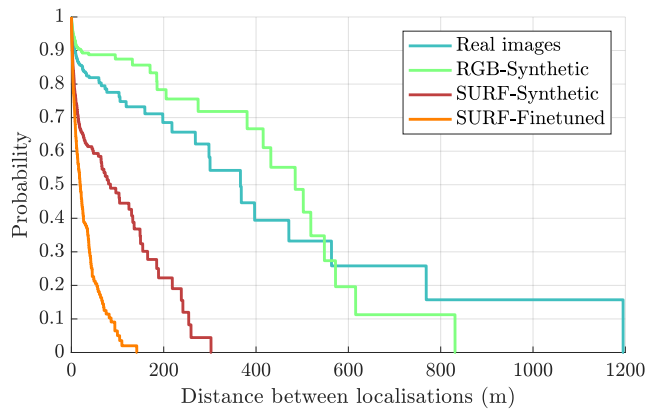

Fig. 9. Probability of travelling a certain distance without localisation against the map (VO-based open-loop or dead-reckoning) when a localisation failure occurs. We observe a large increase in the robustness of day-night localisation when using synthetic images. The map is from daytime, the input images are at night.

## VI. CONCLUSIONS

We have presented a system that yields robust localisation under very adverse conditions. The key to this work is transforming an input image in such a way as to enhance point wise matching to a stored image. This transformation is learnt using a cyclic GAN while explicitly accounting for the attributes feature detection and description stages. We motivated this extra-complexity by showing quantitatively that it is not sufficient to simply learn a transfer function for pixel-wise RGB values - one needs to account for feature detector and descriptor responses. Using modest target training data, which emulates scenarios where mapping is expensive, time-consuming or difficult, the resulting synthetic images consistently improved place recognition and metric localisation compared to baselines. This speaks to not only drastically reducing the cost and inconvenience of mapping under diverse conditions, but also improving the efficacy of the maps produced when used in conjunction with our method. Finally, an architectural nicety is that our system processes the image streams outside of the localisation pipeline, either offline or online, and hence can be used naturally as a front end to many existing systems.

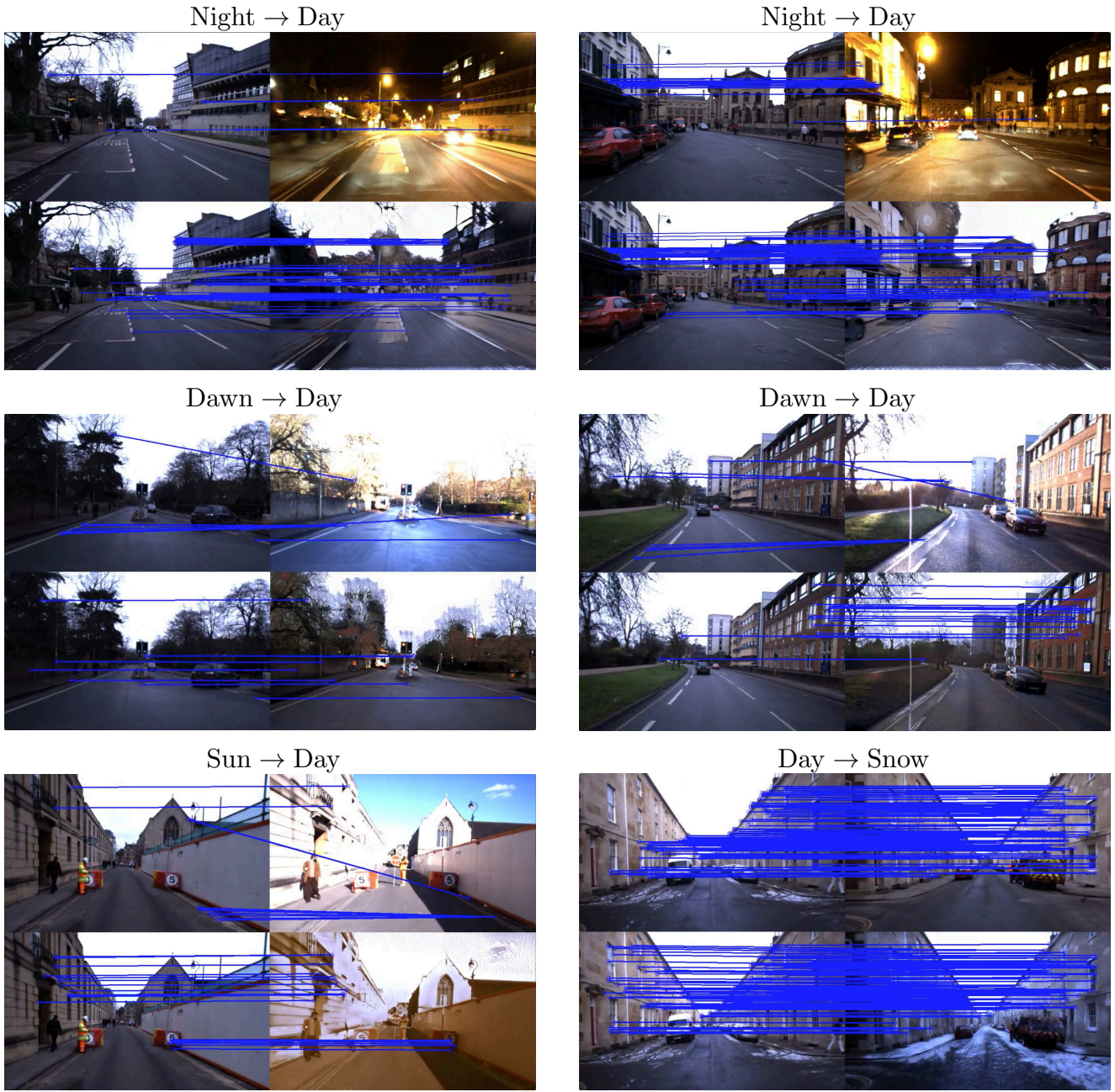

Fig. 10. Example feature-based localisation between different conditions using appearance transfer. In all cases our approach provides a greater number of inlier matches (blue) when using the transformed image compared to matching the original images captured under significantly different conditions. Top left and bottom left represent the real-day map image. Top right represents the original live camera frame. Bottom right represents the live camera frame transformed to match the condition of the map image. These results are obtained using stage-1 of our architecture.

| **Map vs. Traversal** | **REFERENCE** | | | **RGB-only** | | | **SURF** | | | **SURF + finetune** | | |
|---|---|---|---|---|---|---|---|---|---|---|---|---|
| | % | RMSE (m) | RMSE (°) | % | RMSE (m) | RMSE (°) | % | RMSE (m) | RMSE (°) | % | RMSE (m) | RMSE (°) |
| Real Day vs Real Night | 19.5 | 12.74 | 24.06 | — | — | — | — | — | — | — | — | — |
| Real Day vs Synthetic Day | — | — | — | 25 | 3.59 | 11.38 | 51 | 2.43 | 4.06 | 60 | 1.99 | 3.32 |

TABLE I
METRIC LOCALISATION PERFORMANCE FOR DIFFERENT NETWORK ARCHITECTURES

| Map vs. Traversal | REFERENCE | | | SYNTHETIC | | |
|---|---|---|---|---|---|---|
| | % | RMSE (m) | RMSE (°) | % | RMSE (m) | RMSE (°) |
| Day vs Night | 19.5 | 12.74 | 24.06 | 51.0 | 2.43 | 4.06 |
| Day vs Snow | 94.6 | 9.02 | 8.07 | 98.6 | 3.38 | 8.19 |
| Day vs Dawn | 31.9 | 11.59 | 7.04 | 66.8 | 2.55 | 6.30 |
| Day vs Sun | 33.0 | 32.35 | 82.04 | 71.0 | 9.40 | 24.52 |
| Day vs Rain | 27.2 | 11.28 | 9.85 | 58.6 | 2.91 | 7.84 |

TABLE II
METRIC LOCALISATION PERFORMANCE BETWEEN CONDITIONS

A video describing our results can be found at https://www.youtube.com/watch?v=s8XV1Y6opig

## VII. FUTURE WORK

Future work could involve a system that can deal with continuous changes in illumination and weather conditions, rather than binning them into discrete categories, and ideally learning a single generator model in order to accomplish this.

## VIII. ACKNOWLEDGEMENTS

This work was supported by Oxford-Google DeepMind Graduate Scholarships and Programme Grant EP/M019918/1.